\documentclass[conference]{IEEEtran}
\usepackage{amsmath,graphicx}
\usepackage{epsfig}
\usepackage{graphicx}
\usepackage{amsmath}
\usepackage{amssymb}
\usepackage{algorithm}
\usepackage{algorithmic}
\usepackage{enumerate}
\usepackage{multirow}
\usepackage{multicol}
\usepackage{arydshln}
\usepackage{color}
\usepackage{amssymb}
\usepackage[font=small]{caption} 
\usepackage{subcaption}

\def\0{{\mathbf 0}}
\def\1{{\mathbf 1}}

\def\a{{\mathbf a}}

\def\f{{\mathbf f}}

\def\v{{\mathbf v}}

\def\y{{\mathbf y}}

\def\D{{\mathbf D}}

\def\F{{\mathbf F}}

\def\F{{\mathbf F}}

\def\L{{\mathbf L}}
\def\M{{\mathbf M}}

\def\S{{\mathbf S}}

\def\W{{\mathbf W}}

\def\ie{{\textit{i.e.}}}
\def\eg{{\textit{e.g.}}}

\def\cE{{\mathcal E}}

\def\cG{{\mathcal G}}

\def\cO{{\mathcal O}}

\begin{document}

\title{Signal Processing in the Retina: Interpretable Graph Classifier to Predict Ganglion Cell Responses}
%
\author{
\IEEEauthorblockN{Yasaman Parhizkar, Gene Cheung, and Andrew W. Eckford}
\IEEEauthorblockA{Dept. of Electrical Engineering and Computer Science\\ York University, Toronto, Canada}
}

%
%
%

\maketitle
\begin{abstract}
It is a popular hypothesis in neuroscience that ganglion cells in the retina are activated by selectively detecting visual features in an observed scene. 
While ganglion cell firings can be predicted via data-trained deep neural nets, the networks remain indecipherable, thus providing little understanding of the cells' underlying operations.
To extract knowledge from the cell firings, in this paper we learn an interpretable graph-based classifier from data to predict the firings of ganglion cells in response to visual stimuli. 
Specifically, we learn a positive semi-definite (PSD) metric matrix $\M \succeq 0$ that defines Mahalanobis distances between graph nodes (visual events) endowed with pre-computed feature vectors; the computed inter-node distances lead to edge weights and a combinatorial graph that is amenable to binary classification.
Mathematically, we define the  objective of metric matrix $\M$ optimization using a graph adaptation of large margin nearest neighbor (LMNN), which is rewritten as a semi-definite programming (SDP) problem. 
We solve it efficiently via a fast approximation called Gershgorin disc perfect alignment (GDPA) linearization.
The learned metric matrix $\M$ provides interpretability: important features are identified along $\M$'s diagonal, and their mutual relationships are inferred from off-diagonal terms. 
Our fast metric learning framework can be applied to other biological systems with pre-chosen features that require interpretation.
\end{abstract}
\section{Introduction}
\label{sec:intro}

The retina is a thin layer of nerve tissue at the back of the vertebrate eye, which receives images and transmits them as electric pulses through the optic nerve to the brain. 
It consists of several layers of cells: from photoreceptors which detect light and primary colors, to ganglion cells with long axons stemming from the optic nerve \cite{remington2005}.
As such, the retina is an important example of biochemical and neurological signalling, both detecting and preprocessing biological signals before transmitting them to the brain's visual cortex.

In the analysis of retinal signals, one hypothesis conjectures that each ganglion cell type computes one or more specific features of the visual scene through a dedicated neural circuit, which connects it to the photoreceptors \cite{gollisch2010eye}.
This implies that \textit{the brain's downstream regions do not receive a general representation of the image, but instead receive a highly processed set of extracted features}. 
For instance, some types of ganglion cells were found responding to features such as motion in a specific direction, texture motion, and anticipation of a periodic stimulus \cite{gollisch2010eye}. Thus, it is natural to consider whether those features can be extracted from the firing patterns. 

State-of-the-art cell firing prediction algorithms employ deep learning, such as Convolutional Neural Networks (CNN) \cite{mcintosh2016deep, lozano20183d, tanaka2019deep} and Recurrent Neural Networks (RNN) \cite{batty2017multilayer}.
While deep learning models predict well, they are inherently uninterpretable ``black boxes" and fail to reveal biological information about what exactly trigger cell firings.
\textit{In response, the main contribution of this paper is to extract knowledge from the cell firings by learning an interpretable graph-based binary classifier from data.} 
Specifically, we learn a \textit{positive semi-definite} (PSD) metric matrix $\M \succeq 0$ that defines Mahalanobis distances between graph nodes (visual events) endowed with pre-computed feature vectors.
The computed node-to-node distances lead to edge weights and a finite graph that is amenable to clustering and binary classification.

Mathematically, unlike the previous \textit{graph Laplacian regularizer} (GLR) \cite{pang17} used in metric learning \cite{yang2006distance}, we adapt the \textit{large margin nearest neighbor} (LMNN) method \cite{weinberger2009distance} into our graph setting as an objective function, and
formulate a \textit{semi-definite programming} (SDP) problem \cite{luo10} to optimize $\M$. We efficiently solve it via an adoption of a fast approximation called \textit{Gershgorin disc perfect alignment} (GDPA) linearization \cite{yang22}.
The learned $\M$ provides a level of interpretability: features crucial to classification can be identified immediately along $\M$'s diagonal, while their mutual relationships can be directly inferred from off-diagonal terms.

The employed features in this paper are localized spatially, and one set (3D-SIFT) has clear interpretation in terms of local image gradients. 
This can enable the estimation of a ganglion cell's receptive field \cite{MARSHAK2009211} and type \cite{gollisch2010eye} based on the features to which it is most sensitive. 
This estimation is inferred directly from the natural stimulus data without the need for random noise stimulus data \cite{NIPS2007_289dff07} or more complex experiments.
More generally, our fast metric learning framework is applicable to other biological systems with pre-chosen features that require interpretation.

We organize the paper as follows.
In Section\;\ref{sec:retina}, we review previous attempts at understanding the retina's visual code. We discuss the shortcomings of these works and establish the need for interpretability.
In Section\;\ref{sec:graph-classifier}, we describe our proposed model along with the training and validation processes. We delineate how the Mahalanobis distance metric can be trained by minimizing the GLR, then we discuss GLR's limitations.
Section\;\ref{sec:optimize} manifests the main contribution of this paper: adapting LMNN to the graph setting as a new training objective called \textit{graph-based large margin nearest neighbour} (GLMNN) and approximating the solution using GDPA linearization for faster computation.
Experimental results and conclusion are presented in Section\;\ref{sec:results} and \ref{sec:conclude}, respectively.
\vspace{-0.1in}

\section{The Visual Code of the Retina}
\label{sec:retina}

The retina is a complex network of diverse cells intricately connected to each other in smaller local circuits. These circuits may have various roles, from basic enhancements to the eye's visual input to extracting sophisticated visual features \cite{gollisch2010eye}.
Many works investigate the cellular circuits of the retina and their role in extracting specific features from the eye's visual input \cite{demb2001bipolar, berry1999anticipation, olveczky2003segregation, munch2009approach}.
Although these investigations provide thorough insight into the retina's structure and function, their development is slow and difficult.
Another standard approach is the Linear-nonlinear (LN) model \cite{chichilnisky2001simple} which combines a linear spatiotemporal filter with a single static nonlinearity.
These models have been used to describe the retinal responses to artificial stimuli; however, they fail to generalize to natural stimuli \cite{heitman2016testing}.

Deep learning models significantly outperform traditional approaches in predicting retinal responses both to artificial and natural stimuli as demonstrated in \cite{mcintosh2016deep, lozano20183d, tanaka2019deep, batty2017multilayer}.
However, as discussed in \cite{tanaka2019deep}, the opaqueness of these models makes it difficult to gain insight into the retinal cells' collective or individual behavior.
A systematic approach to shed light on the trained deep learning models is proposed in \cite{tanaka2019deep} using dimensionality reduction and modern attribution methods.

The approach in \cite{tanaka2019deep} falls under a category of interpretability techniques called ``post-hoc'' methods \cite{linardatos2020explainable}. 
These methods attempt to interpret a black-box model after it is designed and trained.
Another category is the creation of intrinsically more transparent AI models, including our approach: we employ interpretable components to build a more comprehensible classification system, while maintaining competitive prediction accuracy with state-of-the-art methods.

To draw a parallel between our approach and \cite{tanaka2019deep}, we can compare the attribution scores presented in \cite{tanaka2019deep} to the diagonal entries of our computed metric matrix \(\M^*\) in our method, which can be further used for feature selection.
However, there is no equivalence to the non-diagonal entries of \(\M^*\) in \cite{tanaka2019deep}---the relationship between features themselves are left unexplored.
Moreover, our model provides an intuitive map of all datapoints; the distance of a newly introduced datapoint to known datapoints can be easily inferred and understood.

\section{Graph Metric Learning and GLR optimization}
\label{sec:graph-classifier}
We first discuss metric learning on graphs and juxtapose our model with previously developed approaches.
Then, we describe the dataset used in our experiment. 
Lastly, the training and validation processes are described.
We show how GLR can serve as the training objective, albeit with limitations.

\subsection{Metric Learning on Graphs}

Distance metric learning---computation of feature distance between two items endowed with feature vectors---is a popular machine learning sub-topic, encompassing different notions of distance and corresponding algorithms, such as supervised global/local metric learning, unsupervised methods, SVN methods and kernel methods \cite{yang2006distance}.
Notably, for binary classification authors in \cite{weinberger2009distance} optimized the \textit{Mahalanobis distance}, defined as 
\begin{align}
d_{i,j} \triangleq (\f_i - \f_j)^\top \M (\f_i - \f_j) ,
\label{equ:Mahalanobis_def}
\end{align}
where \(\M\) is the PSD metric matrix, and \(\f_i, \f_j\) are the feature vectors of data points \(i, j\), respectively.
This distance is optimized by minimizing the \textit{large margin nearest neighbors} (LMNN) objective given data. 
In a nutshell, via metric matrix $\M$, LMNN minimizes
the distances between same-label pairs while maximizing the distances between different-label pairs, but no further than a large margin.

Our metric matrix optimization objective is similar to one in \cite{weinberger2009distance} in that it defines Mahalanobis distance as the chosen metric for same- / different-label pairs. 
However, we adapt the LMNN objective to our constructed graph model instead of a multi-dimensional space. 
Specifically, in our model datapoints are nodes of a connected graph, and the node labels collectively constitute a \textit{graph signal} on the underlying graph kernel \cite{ortega18ieee,cheung18}.
We call our newly adapted objective \textit{Graph-based Large Margin Nearest Neighbor} (GLMNN).

\begin{figure}[h]
\centering
\captionsetup{justification=centering,margin=1cm}
\vspace{-0.1in}
\begin{subfigure}{.48\textwidth}
  \centering
  \includegraphics[width=\linewidth]{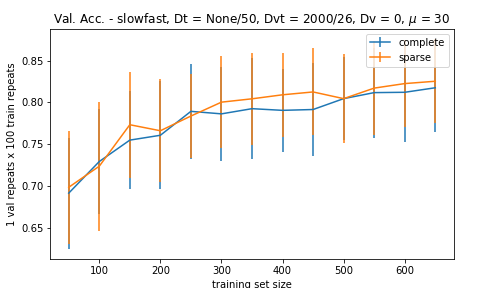}
  \caption{Accuracy}
  \label{fig:complete_v_sparse2-a}
\end{subfigure}\\
\begin{subfigure}{.48\textwidth}
  \centering
  \includegraphics[width=\linewidth]{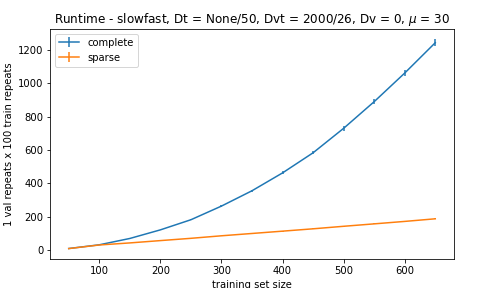}
  \caption{Runtime}
  \label{fig:complete_v_sparse2-b}
\end{subfigure}
\caption{Comparison between graph-based classifiers for a \textit{complete} graph and a \textit{sparse} graph using the GLR objective, as function of training dataset size, in (a) prediction accuracy and (b) runtime.}
\vspace{-0.1in}
\label{fig:complete_v_sparse2}
\end{figure}

One key advantage of using a graph in the metric learning context is that we can control the complexity of the training process by strategically choosing \textit{sparse} connections between nodes.
For instance, Fig.\;\ref{fig:complete_v_sparse2} compares the accuracy and runtime of the model in two cases: when all graph edges are present, and when the maximum degree of nodes is fixed at a constant number. 
Refer to Section\;\ref{sec:experiment_setup} for more information about how the sparse edges are selected.
We observe that runtime can be greatly reduced while prediction accuracy is maintained.
Each edge in the graph corresponds with one term in the optimization objective. 
As a result, by strategically removing edges from the graph, we can simplify the objective without significantly degrading performance.

In \cite{hu2020feature} and \cite{li2022fast}, the authors trained PSD metric matrix $\M$ by minimizing the Mahalanobis distance to restore noisy graph signals and classify entries with binary labels, respectively.
The training objective is a quantity called \textit{Graph Laplacian Regularizer} (GLR) \cite{pang17} instead of LMNN.
However, for binary classification GLR does not efficiently utilize all available information in the training data; we elaborate this point in the sequel.
Lastly, \cite{yang22} proposed the \textit{Gershgorin Disc Perfect Alignment} (GDPA) linearization technique to reduce the complexity of optimizing a matrix variable with a PSD cone constraint.
A faster version of GDPA is also used in \cite{li2022fast}.
We employ GDPA linearization to speed up the minimization of GLMNN during training.

\subsection{The Data}
\label{sec:the_data}

\begin{figure}[t!]
\centering
\captionsetup{justification=centering,margin=0.1cm}
\includegraphics[width=0.45\textwidth]{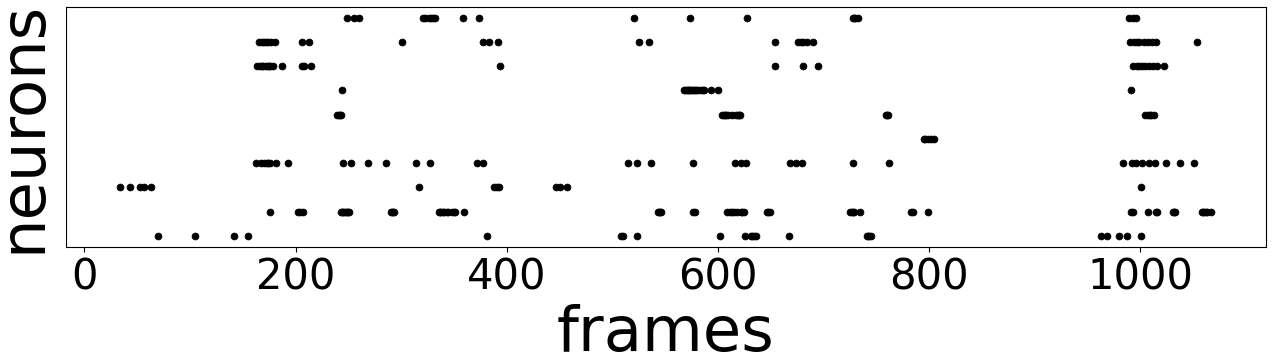}
\caption{Ganglion cells' spikes through time in response to a movie.}
\label{fig:spikes}
\end{figure}

We use the experimental data from \cite{marre2017multi}, where individual ganglion cells' spikes are recorded through time while a naturalistic movie was being played in front of an extracted salamander's retina (see Fig.\;\ref{fig:spikes}). 
The movie has $1141$ frames, each of size $120 \times 200$ pixels.
Each frame was displayed for $1/60$ seconds. 
The time axis was separated into bins of $1/60$ seconds synced with the frames.
Each row of the figure shows a specific cell's spikes, where dots represent spikes.
The spike signals are binary because a ganglion cell can either fire or stay silent at a given moment; hence, a binary number can describe its state.
The brain receives numerous binary signals from many ganglion cells and analyzes all signals together in the process of visual perception.
The responses of $113$ cells to $297$ repeated playbacks of the same movie were recorded. 
As a result, the whole dataset is stored as two tensors: one tensor describing the stimulus shaped as $1141 \times 120 \times 200$, and one describing the cell responses shaped as $113 \times 297 \times 1141$.

Individual cells fire in only about $3\%$ of the time bins.
To create more balance between firings and idleness, we distinguish the ``all-silent'' state, when no neuron is firing, from any other state when at least one neuron is active. 
The ``all-silent'' state frequently appears in retinal recordings, and it is well known in the literature \cite{tkavcik2014searching}.
Hence, we group all ganglion cells so that a time bin's label in the group response is $1$ (black) if \textit{any} ganglion cell in a group spikes during that bin, and $-1$ (white) otherwise. 
Labels are denoted by $y_i \in \{-1, 1\}$ for $i \in \{1, \ldots, 1141\}$.
Thus, the group response tensor is shaped as $1 \times 297 \times 1141$.

Further, we consider a batch of $U$ frames immediately preceding the time bin's own frame to predict the label.
The number of frames in a batch depends on the utilized feature extraction algorithm. In our experiments, we used batches of $U = 32$ and $U = 42$ frames.
Features for each batch are extracted using methods such as pre-trained CNNs' filters or hand-crafted feature sets like SIFT \cite{lowe2004distinctive}. 
The resulting $K$ features extracted from each batch $i$ are denoted by vector $\f_i \in \mathbb{R}^K$. 
In summary, our dataset consists of \((\f_i, y_i)\) pairs describing a feature vector extracted from a frame batch and its corresponding label. 
A single \((\f_i, y_i)\) pair is called a datapoint.

\subsection{Graph Construction \& GLR Optimization}
First, we split the data into training and validation sets.
Datapoints are selected randomly for the sets while ensuring that both sets consist of half $1$-labeled datapoints and half $-1$-labeled datapoints.
We then construct a \textit{similarity graph} $\cG_t$ with $N$ nodes representing $N$ labeled training datapoints, each with a binary label $y_i$ and a feature vector $\f_i$.
We define the weight $w_{i,j}$ of an edge connecting nodes \(i\) and \(j\) using an exponential kernel and Mahalanobis distance:
\vspace{-0.05in}
\begin{align}
w_{i,j} = \exp\left( - d_{i,j} \right) ,
\label{equ:edge-weights}
\end{align}
\noindent where \(d_{i,j}\) is defined in Equation\;\eqref{equ:Mahalanobis_def} which depends on $\M \in \mathbb{R}^{K \times K}$. \(\M\) is a \textit{metric matrix} learned through a training process.
Moreover, $\M \succeq 0$ is PSD to ensure $d_{i,j} \geq 0$.
The edge set $\{w_{i,j}\}$ connecting $N$ nodes thus defines an \textit{adjacency matrix} $\W_t \in \mathbb{R}^{N \times N}$ that specifies graph $\cG_t$. 
The training graph is illustrated in Fig.\;\ref{fig:val-graph} with blue nodes and edges. 
The red nodes represent the validation datapoints, which are added to the graph after training. 
The goal of training is to find an optimal metric matrix \(\M\) such that edges connecting two same-label nodes have noticeably higher weights than edges between opposing-label nodes.
If successful, same-label nodes would form local-neighborhood \textit{clusters} as illustrated, and thus the constructed graph becomes amenable to classification.

Let the training objective $Q(\M)$ be the GLR (as in \cite{hu2020feature}) on known label signal $\y_t \in \{-1, 1\}^{N}$ plus a trace term to ensure that edge weights do not all tend to zero:
\begin{align}
\min_{\M \succeq 0} \; Q(\M) = \y_t^\top \L_t(\M) \y_t + \mu \, \text{tr}(\M)
\label{equ:objective}
\end{align}
\noindent 
where $\L_t \triangleq \text{diag}(\W_t \1) - \W_t$ is the graph Laplacian matrix, $\mu > 0$ is a chosen parameter,
$\1$ is an all-one vector of suitable length, and $\text{diag}(\v)$ returns a diagonal matrix with diagonal terms $\v$.
Note that Laplacian $\L_t$ (by extension graph $\cG_t$) is a function of $\M$, since $\M$ defines distances $d_{i,j}$, which in turn determines edge weights $w_{i,j}$ and $\W_t$. 
The trace term upper bounds the eigenvalues of \(\M\) which, in turn, ensures that the distance between two finite feature vectors remains finite.

We can expand the objective $Q(\M)$ in \eqref{equ:objective} and substitute the definition of Mahalanobis distance from \eqref{equ:Mahalanobis_def} to arrive at a more clear formulation,

\vspace{-0.05in}
\begin{small}
\begin{align}
\min_{\M \succeq 0}
\sum_{(i,j) \in \cE} \underbrace{\exp \left( -(\f_i - \f_j)^\top \M (\f_i - \f_j) \right)}_{w_{i,j}} (y_i - y_j)^2 + \mu \, \text{tr}(\M)
\end{align}
\end{small}\noindent 
where the summation is the connected node-pair expansion of GLR \cite{pang17} $\y^\top \L(\M) \y$, and $\y \in \{-1,1\}^N$ is the aforementioned known label signal. (We drop the subscript $t$ in the sequel for notational convenience.)
When $y_i = y_j$---\ie, nodes $i$ and $j$ have the same labels---each term $w_{i,j}(y_i - y_j)^2 = 0$ regardless of the value of $w_{i,j}$. 
However, to properly train the metric matrix $\M$, the knowledge that {\em same-label pairs should be close in feature distance} can be useful; specifically, $d_{i,j} = (\f_i-\f_j)^\top \M (\f_i-\f_j) \geq 0$ for same-label pair $(i,j)$ should be small.

\subsection{Estimating Labels from Graph}

\begin{figure}[h]
\centering
\captionsetup{justification=centering,margin=0.1cm}
\includegraphics[width=0.45\textwidth]{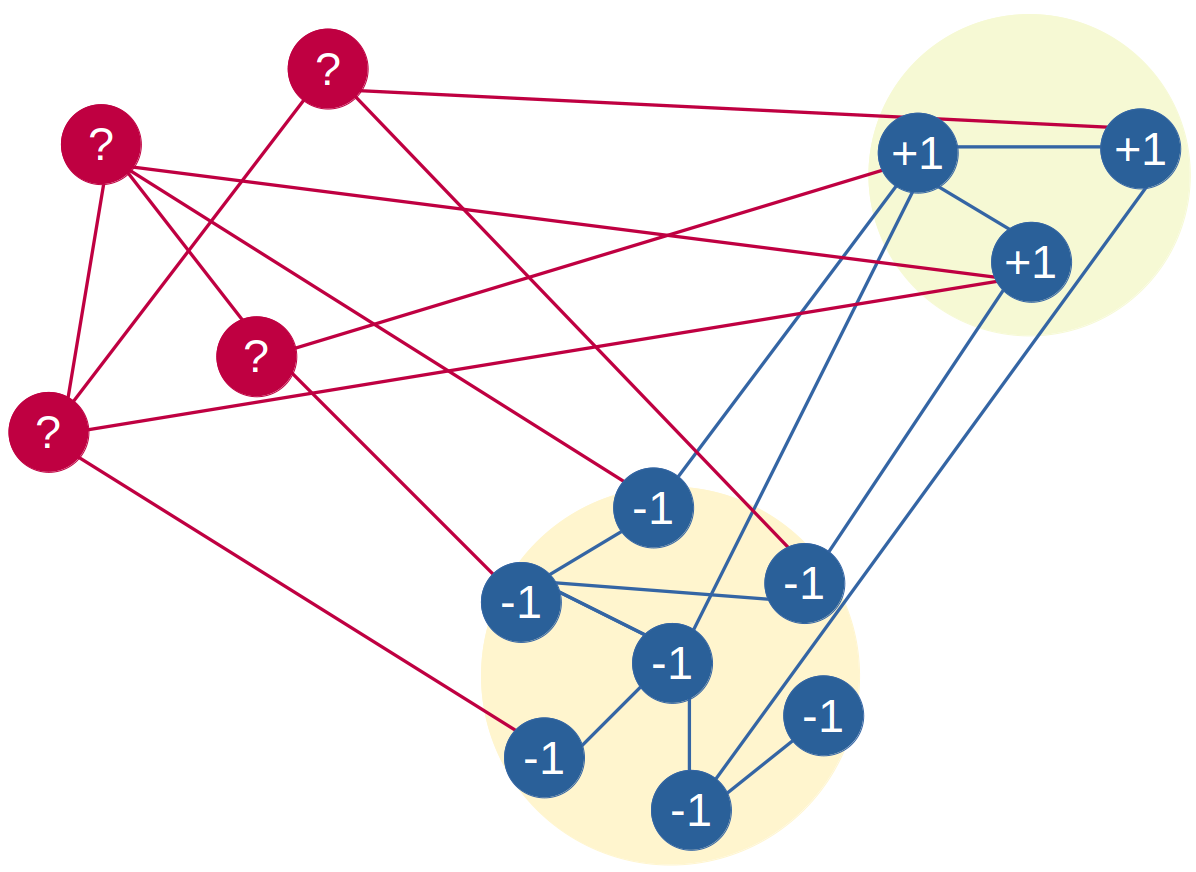}
\caption{Training (blue) and validation (red) nodes in the similarity graph.}
\label{fig:val-graph}
\end{figure}

After training, we are left with a similarity graph and an optimized metric matrix.
To estimate missing labels, we represent each one of $M$ validation visual events by an additional node (red nodes in Fig.\;\ref{fig:val-graph}).
We connect each validation node to training nodes, making sure that at least one training node from each label is connected. 
Then, we minimize the GLR of this expanded graph $\cG$ of $N + M$ nodes:
\vspace{-0.05in}
\begin{align}
\min_{\y_v} \; \y^\top \L \y ,
\label{equ:estimation}
\end{align}
\noindent 
where $\L \in \mathbb{R}^{(N+M) \times (N+M)}$ is the expanded graph's Laplacian, and $\y = \left[ \y_t^\top \; \y_v^\top \right]^\top$ consists of both training and validation labels.

After minimization, we hard-threshold the entries of $\y_v^*$ to obtain final visual event label predictions: 
\begin{align}
(\hat{\y}_v)_i = \left\{ \begin{array}{ll}
1 & \mbox{if}~ (\y_v^*)_i > 0 \\
-1 & \mbox{o.w.}
\end{array} \right. .
\label{equ:prediction}
\end{align}
If \((\hat{y}_v)_i = 1\), then we predict a spike will occur, while if  \((\hat{y}_v)_i = -1\), then we predict a spike will not occur.

\section{GLMNN Optimization and GDPA Approximation}
\label{sec:optimize}

We present a new training objective---a graph adaptation of LMNN---and reformulate the problem as a semi-definite program (SDP). 
Then, we solve the posed SDP efficiently via GDPA linearization \cite{yang22}.

\subsection{Graph-based Large Margin Nearest Neighbor Objective}

As previously discussed, when GLR is used as the training objective, same-label node pairs are not used to optimize metric matrix $\M$. 
However, enforcing large edge weights (small Mahalanobis distances) between same-label node pairs can inform a better similarity graph. 
This motivates an alternative training objective. 

For binary classification, we propose a new objective called \textit{graph-based large margin nearest neighbor} (GLMNN) focusing on same- \textit{and} opposing-label pair distances:
\begin{align}
\min_{\M \succeq 0} & \sum_{(i,j) \in \cE | y_i = y_j} d_{i,j}(\M)
\nonumber \\
&+ \rho \!\!\!\!\! \sum_{(i,j), (i,l) \in \cE | y_i = y_j = -y_l}
\!\!\!\!\!\!\!\! \left[ d_{i,j}(\M) + \gamma - d_{i,l}(\M) \right]_+
\label{eq:GLMNN}
\end{align}
where $d_{i,j}(\M) = (\f_i-\f_j)^\top \M (\f_i - \f_j)$, $\gamma > 0$ is a parameter, and $[a]_+$ returns $a$ if $a \geq 0$ and $0$ otherwise.
The first sum is an aggregate of distances $d_{i,j}(\M)$ between same-label node pairs connected by edges $(i,j) \in \cE$. 
The second sum means that each distance $d_{i,l}(\M)$ between opposing-label pair $(i,l)$ would be penalized \textit{only if} it is within $\gamma$ of a same-label pair distance $d_{i,j}(\M)$. 
Though the idea of trading off same- and opposing-label pair distances originates from LMNN \cite{weinberger2009distance}, 
GLMNN adapts LMNN into a graph context, where same- / opposing-label pairs $(i,j)$ are specified as graph nodes $i$ and $j$ connected by edges $\cE$ in a graph $\cG$. 
An intentionally sparsely constructed graph $\cG$ would thus lead to lower complexity when computing objective \eqref{eq:GLMNN} due to fewer edges in $\cE$. 

Towards efficient optimization, we first rewrite the objective \eqref{eq:GLMNN} as follows.
First, we rewrite each distance term $d_{i,j}(\M)$ in the two sums as
\begin{align}
\nonumber
d_{i,j}(\M) &= (\f_i - \f_j)^\top \M (\f_i - \f_j)\\ 
&= 
\text{tr}\left( (\f_i - \f_j)^\top \M (\f_i - \f_j) \right) 
\nonumber \\
&= \text{tr} \left( \M (\f_i - \f_j) (\f_i - \f_j)^\top \right) 
\nonumber \\
&= \text{tr} (\M \F_{i,j} )
\end{align}
where $\F_{i,j} \triangleq (\f_i - \f_j) (\f_i - \f_j)^\top$.
Second, to linearize the non-negativity function $[\a]_+$, we define non-negative auxiliary variable $\delta_{i,j,l}$ that is an upper bound on $d_{i,j}(\M) + \gamma - d_{i,l}(\M)$. 
Hence, minimizing $\delta_{i,j,l}$ is equivalent to minimizing $[d_{i,j}(\M) + \gamma - d_{i,l}(\M)]_+$.
The optimization \eqref{eq:GLMNN} can now be rewritten as
\begin{align}
\min_{\{\delta_{i,j,l}\}, \M \succeq 0} & \sum_{(i,j) \in \cE | y_i = y_j} \text{tr}(\M \F_{i,j}) + \rho \!\!\!\!\!\!\!\! \sum_{(i,j), (i,l) \in \cE | y_i = y_j = -y_l} \!\!\!\!\!\!\!\! \delta_{i,j,l} 
\nonumber \\
\mbox{s.t.} & ~~~ \delta_{i,j,l} \geq \text{tr}(\M \F_{i,j}) + \gamma - \text{tr}(\M \F_{i,l}) 
\nonumber \\
& ~~~ \delta_{i,j,l} \geq 0
\label{eq:obj2}
\end{align}
where the variables are $\{\delta_{i,j,l}\}$ and $\M$. 
The problem is a \textit{semi-definite programming} (SDP) problem with linear objective and linear constraints plus a PSD cone constraint $\M \succeq 0$ \cite{luo10}.

While SDP problems can be solved in polynomial time using an off-the-shelf SDP solver, the worst-case complexity $\cO(K^3)$ is still high.
Next, we adopt a fast approximation algorithm that addresses convex optimization problems with a PSD cone constraint called \textit{GDPA linearization} \cite{yang22}.

\subsection{GDPA Linearization Algorithm}
\label{subsec:GDPAL}

\subsubsection{Replacing PSD Cone Constraint}

The basic idea of GDPA linearization is to replace the PSD cone constraint with a set of linear constraints per iteration, so that together with the linear objective and other linear constraints, the SDP problem simplifies to a \textit{linear program} (LP), solvable using a state-of-the-art LP solver\footnote{A representative state-of-the-art general LP solver is \cite{jiang20}, which has complexity $\mathcal{O}(K^{2.055})$.} per iteration. 
GDPA linearization is based on a well-known linear algebra theorem called the \textit{Gershgorin Circle Theorem} (GCT) \cite{varga04}, a version of which states that each real eigenvalue $\lambda$ of a real symmetric matrix $\M$ resides inside at least one \textit{Gershgorin disc} $\Psi_i$ corresponding to row $i$, with center $c_i \triangleq M_{i,i}$ and radius $r_i \triangleq \sum_{j \neq i} |M_{i,j}|$, \ie, $\exists i, \mbox{s.t.} ~ c_i - r_i \leq \lambda \leq c_i + r_i$. 
The corollary is that the smallest eigenvalue $\lambda_{\min}$ is lower-bounded by the smallest disc left-end $\lambda_{\min}^-$, \ie, 
\begin{align}
\lambda_{\min}^-(\M) \triangleq \min_i M_{i,i} - \sum_{j\neq i} |M_{i,j}| \leq \lambda_{\min}(\M) .
\end{align}
Thus, to ensure $\M \succeq 0$, one can enforce linear constraints $\lambda_{\min}^-(\M) \geq 0$, \ie, all disc left-ends are at least zero.

However, because the lower bound $\lambda_{\min}^-(\M) \leq \lambda_{\min}(\M)$ is often loose, enforcing $\lambda_{\min}^-(\M) \geq 0$ directly would mean we are overly restricting the search space, not allowing feasible solutions $\M'$ where $\lambda_{\min}^-(\M') < 0 \leq \lambda_{\min}(\M')$.

Knowing that a \textit{similarity transform} $\S \M \S^{-1}$ of $\M$ shares the same set of eigenvalues (assuming $\S$ is invertible), we first replace constraint $\lambda_{\min}^-(\M) \geq 0$ with $\lambda_{\min}^-(\S\M\S^{-1}) \geq 0$.
Assuming a diagonal $\S$, the $N$ disc left-end constraints are
\begin{align}
M_{i,i} - \sum_{j | j\neq i} \left| \frac{s_i M_{i,j}}{s_j} \right| \geq 0, ~~\forall i .
\label{eq:GDPA_linConst}
\end{align}

Second, we can select diagonal $\S$ so that the lower-bound gap can be eliminated for a given $\M$, \ie, $\lambda_{\min}^-(\S \M \S^{-1}) = \lambda_{\min}(\S \M \S^{-1}) = \lambda_{\min}(\M)$, as follows.
First, we recall that a recent linear algebra theorem called \textit{Gershgorin Disc Perfect Alignment} (GDPA) \cite{yang22} states that $\lambda_{\min}^-(\S \M \S^{-1}) = \lambda_{\min}(\M)$ for a diagonal $\S$ if
\begin{enumerate}
\item $s_i = 1/v_i, \forall i$ and $\v$ is the first eigenvector of $\M$, and
\item $\M$ is a Laplacian matrix to a \textit{balanced} signed graph.
\end{enumerate}
By the \textit{Cartwright-Harary Theorem} (CHT) \cite{cartwright56}, a  signed graph (\ie, a graph with both positive and negative edges) is balanced if and only if nodes can be colored into two colors (\eg, blue and red), such that positive / negative edges connect nodes of the same / different colors.
Given GDPA, we can design a fast approximation algorithm to \eqref{eq:obj2} as follows.

\subsubsection{Algorithm Design}

Assume that at iteration $\tau-1$, a feasible solution $\M^\tau \succeq 0$ to \eqref{eq:obj2} has been computed, where $\M^\tau$ is a Laplacian to a balanced signed graph $\cG^s$, \ie, nodes in $\cG^s$ can be assigned colors blue and red, so that edges connecting same- / different-color pairs $(i,j)$ have positive / negative weights $w_{i,j}$ (denoted by $-M_{i,j}^\tau$, since by definition Laplacian $\M^\tau \triangleq \D^\tau - \W^\tau + \text{diag}(\W^\tau)$). 
We first compute first eigenvector $\v$ for $\M^\tau$ using a fast extreme eigenvector computation algorithm such as LOBPCG \cite{knyazev01}, using which we define $s_i = 1/v_i, \forall i$. 
Linear constraints \eqref{eq:GDPA_linConst} that ensure solution $\M \succeq 0$ are now well defined. 

At iteration $\tau$, we optimize diagonal terms $\{M_{l,l}\}_{\forall l}$ plus one row / column $i$ of $\M$ at a time: 
for each row / column $i$, we optimize $\{M_{i,j}, M_{j,i}\}_{\forall j \neq i}$ twice, each time assuming node $i$ is colored either red or blue. 
The assumed color implies positive / negative signs for $M_{i,j}$'s and $M_{j,i}$'s to maintain graph balance to other colored nodes $j \neq i$, which we can enforce using additional linear constraints in a LP.
To illustrate, denote the set of all blue nodes of  $\cG^s$ at iteration $\tau$ by $\mathcal{N}_{b,\tau}$ and the set of all red nodes of  $\cG^s$ at iteration $\tau$ by $\mathcal{N}_{r,\tau}$.
Then, if it is assumed that $i \in \mathcal{N}_{b,\tau}$, the following linear constraints are imposed,
\begin{align}
\left\{ 
\begin{array}{ll}
    M_{i,j} = M_{j,i} \leq 0 & \text{if} \; j \in \mathcal{N}_{b,\tau}\\
    M_{i,j} = M_{j,i} \ge 0 & \text{if} \; j \in \mathcal{N}_{r,\tau}
\end{array}
\right\}
\end{align}
Conversely, if it is assumed that  $i \in \mathcal{N}_{r,\tau}$, the following linear constraints are enforced instead,
\begin{align}
\left\{ 
\begin{array}{ll}
    M_{i,j} = M_{j,i} \leq 0 & \text{if} \; j \in \mathcal{N}_{r,\tau}\\
    M_{i,j} = M_{j,i} \ge 0 & \text{if} \; j \in \mathcal{N}_{b,\tau}
\end{array}
\right\}
\end{align}

We select the smaller of the two objective values corresponding to the two assumed colors as the color for node $i$. 
We optimize all rows / columns $i$ in this manner in turn till convergence, resulting in optimal solution $\M^{\tau+1}$ for iteration $\tau$.
A new first eigenvector $\v$ for obtained solution $\M^{\tau+1}$ is computed again using LOBPCG, and $\{s_i\}$ for linear constraints \eqref{eq:GDPA_linConst} for the next iteration are defined. 
The algorithm continues till solution $\M^\tau$ converges.

\section{Experiments}
\label{sec:results}
\subsection{Experiment Setup}
\label{sec:experiment_setup}

We first extracted visual features from a target fish movie \cite{marre2017multi} using three methods. 
The first is the output of the first layer of a pre-trained CNN called ``slowfast-r50'' \cite{fan2021pytorchvideol, feichtenhofer2019slowfast}.
The second is the output of the second layer of another pre-trained CNN called ``SOE-Net'' \cite{Hadji2017, soe_net}.
The Slowfast network was trained on an action recognition task in videos while SOE-Net was developed for simultaneous audio-video texture analysis.
The third is a 3D version of the well-known \textit{scale-invariant feature transform} (SIFT) \cite{brister, bristercode}. 
The Slowfast and 3D-SIFT algorithms both use batches of 32 frames while SOE-Net uses batches of 42 frames.
In addition, we used SIFT features extracted from the MNIST dataset \cite{deng2012mnist}, consisted of images of handwritten digits from 0 to 9 and their labels.
Since the neuron dataset contains binary labels ($-1$ and $+1$), we separated only two digits of the MNIST set (0 and 1) to better match our target dataset. 

We compared our algorithm against five representative state-of-the-art classification schemes: CNN as implemented in \cite{mcintosh2016deep, lozano20183d, tanaka2019deep}, RNN similar to \cite{batty2017multilayer}, XGBoost \cite{azmi2020overview} (a Python implementation of gradient boosting decision trees (GBDT) \cite{friedman2001greedy}), logistic regression (LR) \cite{10.1001/jama.2016.7653}, and K nearest neighbors (kNN) \cite{1053964}.
All algorithms were trained on the same features.

After feature extraction, we connected the training and validation nodes as follows.
For each set, we randomly selected half of the nodes from the $1$-labeled bins and the other half from $-1$-labeled nodes; this ensured that the data sets were evenly distributed between $1$ and $-1$ labels.
Edges between two training nodes or two validation nodes were constructed based on their temporal proximity; for example, to connect a node corresponding to time interval \(t\) to five other nodes, we chose nodes corresponding to time intervals closest to \(t\), three of them before \(t\) and two after \(t\).
Edges between a training node and a validation node were selected with a similar strategy; the only difference was that we ensured half the connected nodes were labeled $1$ and half are labeled $-1$.
The maximum number of edges between two training nodes was set at parameter \(D_t\), between two validation nodes was set at \(D_v\), and between a training and a validation node was set at \(D_{vt}\).
Thus, the maximum degree of a training node was \(D_t + D_{vt}\), and the maximum degree of a validation node was \(D_v + 2D_{vt}\).

After constructing a graph connecting all training and validation nodes, the metric matrix was optimized on the graph as described in Section\;\ref{sec:optimize}.
We tested three metric matrix optimization algorithms. 
The first was the GLR objective \eqref{equ:objective} optimized via gradient descent.
The second was the GLMNN objective \eqref{eq:obj2} optimized via an off-the-shelf SDP solver called SeDuMi \cite{strum2001matlab}.
The third optimized the same GLMNN objective \eqref{eq:obj2} via our proposed GDPA linearization algorithm described in Section\;\ref{subsec:GDPAL}.
To randomize our experiments, we repeated each experiment with $P_t$ different training sets and $P_v$ different validation sets, meaning that each point in the following plots is the average result of $P_t P_v$ trials. The values of $P_t$ and $P_v$ are reported on the y-axis of each plot.

\subsection{Experiment Results}
\label{sec:exp_results}

An important advantage of a graph-based classifier is the use of computed similarity edges connecting validation nodes based on available per-node features to estimate their labels, instead of computing labels for each validation node individually.
This property is important for semi-supervised learning, where the training set is small compared to the test set.
We focus on this setting in our experiments.

\subsubsection{Interpretability vs. Performance in Features}

\begin{figure}[h]
\centering
\captionsetup{justification=centering,margin=0.1cm}
\includegraphics[width=0.48\textwidth]{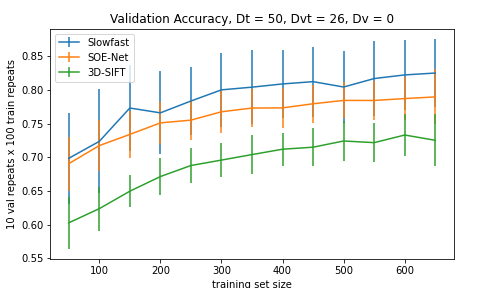}
\caption{Performance of GLR minimization with three different feature extraction algorithms.}
\label{fig:glr-all-features}
\end{figure}

The validation accuracy obtained by minimizing the GLR objective \eqref{equ:objective} are shown in Fig.\;\ref{fig:glr-all-features}.
Each curve illustrates the performance using one of the three feature extraction algorithms described in Section\;\ref{sec:experiment_setup}, all extracted from the fish movie \cite{marre2017multi}.
We observe that Slowfast and SOE-Net achieved a higher accuracy than 3D-SIFT; however, they are less interpretable since their filters are optimized in a purely data-driven manner without clear interpretation.
This illustrates one tradeoff between interpretability and performance. 

\subsubsection{Prediction Accuracy}

\begin{figure}[h]
\centering
\includegraphics[width=0.48\textwidth]{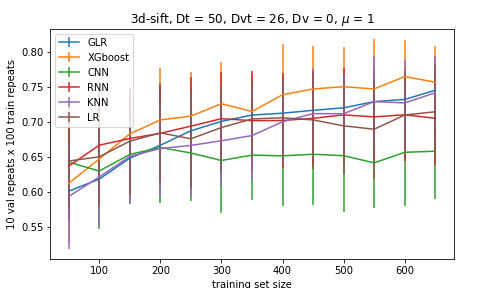}
\caption{Validation accuracy of GLR compared to the state-of-the-art: XGBoost \cite{azmi2020overview}, Convolutional Neural Network (CNN) \cite{mcintosh2016deep, lozano20183d, tanaka2019deep}, Recurrent Neural Network (RNN) \cite{batty2017multilayer}, K-Nearest Neighbors (KNN)\cite{1053964} , and Logistic Regression (LR) \cite{10.1001/jama.2016.7653}.}
\label{fig:factobj1-bench2}
\end{figure}

The prediction accuracy of our graph-based classifier is analyzed jointly by Fig. 5 and 6.
We compared GLR against four other methods in Fig.\;\ref{fig:factobj1-bench2}.
We observe that GLR's performance is on par with the benchmarks.
This shows that a graph-based classifier inherently suffers no performance disadvantage against its competitors.

\begin{figure}[h]
\centering
\captionsetup{justification=centering,margin=0.1cm}
\includegraphics[width=0.48\textwidth]{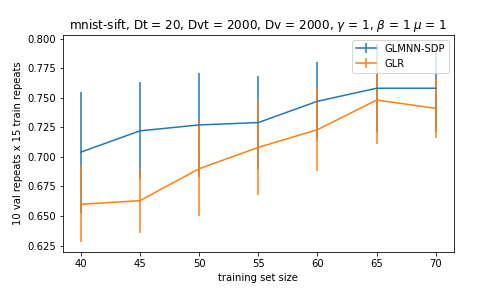}
\caption{Validation accuracy of GLMNN vs. GLR}
\label{fig:glr_lmnn}
\end{figure}

As discussed in Section\;\ref{sec:optimize}, GLMNN utilizes features of same-label pairs to further optimize the metric matrix $\M$. 
However, minimizing GLMNN using an off-the-shelf SDP solver such as SeDuMi is time-consuming.
Thus, we tested SeDuMi only for smaller datasets than in Fig.\;\ref{fig:factobj1-bench2}.
A comparison is made between GLMNN minimization using SeDuMi and GLR in Fig.\;\ref{fig:glr_lmnn}.
We see that GLMNN noticeably outperformed GLR in classification performance.
Further, we observe that the difference between GLMNN and GLR decreased as the training set became larger. One explanation is that the extra information (\ie, distances between same-label pairs) used in GLMNN is more critical for smaller sets where the training data does not provide sufficient information.

\subsubsection{Speed}

\begin{figure}[h]
\centering
\captionsetup{justification=centering,margin=0.1cm}
\includegraphics[width=0.48\textwidth]{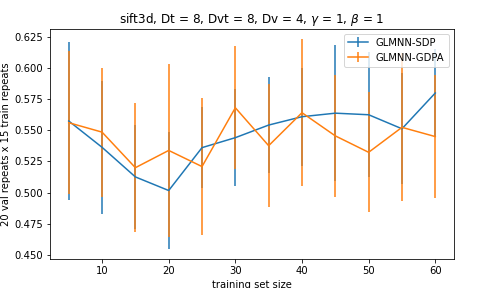}
\caption{Comparison between obtained validation accuracies by SDP and GDPA}
\label{fig:sdp_gdpa_val_acc}
\end{figure}

\begin{figure}[h]
\centering
\captionsetup{justification=centering,margin=0.1cm}
\includegraphics[width=0.48\textwidth]{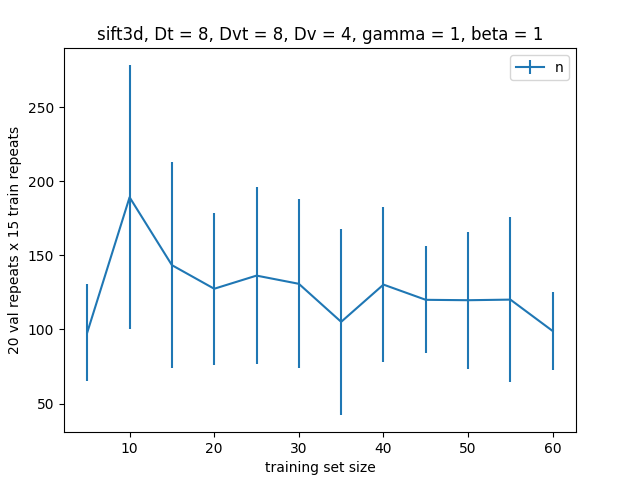}
\caption{The number of times a linear program was run during GDPA optimization.}
\label{fig:num_LPs}
\end{figure}

To speed up GLMNN optimization, GDPA linearization algorithm was used.
The effectiveness of GDPA is illustrated in Fig.\;\ref{fig:sdp_gdpa_val_acc} and \ref{fig:num_LPs}.
Fig.\;\ref{fig:sdp_gdpa_val_acc} compares the obtained accuracy by GLMNN minimization with an SDP solver and GDPA linearization.
We observe that, though GDPA provided only approximate solutions, its performance was reasonably close to an SDP solver as shown in Fig.\;\ref{fig:sdp_gdpa_val_acc}.

On the other hand, GDPA greatly reduces algorithm runtime for large datasets.
A general SDP solver has time complexity \(\cO(K^3)\).
GDPA approximates SDP via a number of iterative linear programs. 
Each linear program requires roughly \(\cO(K^2)\) to compute. 
As observed in Fig.\;\ref{fig:num_LPs}, the number of required linear programs in GDPA till convergence stays roughly constant.
Thus, GDPA reduces time complexity to \(\cO(K^2)\).
The precision of the GDPA algorithm is controlled by the definition of convergence.

\subsubsection{Interpretability}

The optimized metric matrix \(\M^*\) enables  a level of interpretability by depicting the relevance of features to the classification. 
Diagonal elements of \(\M^*\) show the singular contribution of each feature to the distance, while off-diagonal elements show the relevance of features to each other.
It is observed in our experiments that \(\M^*\) tends to have a few large elements, while all the other elements are small. 
This means that only a few features contribute substantially to the classification task, which provides a way for feature selection.

\begin{figure}[h]
\centering
\captionsetup{justification=centering,margin=0.1cm}
\includegraphics[width=0.48\textwidth]{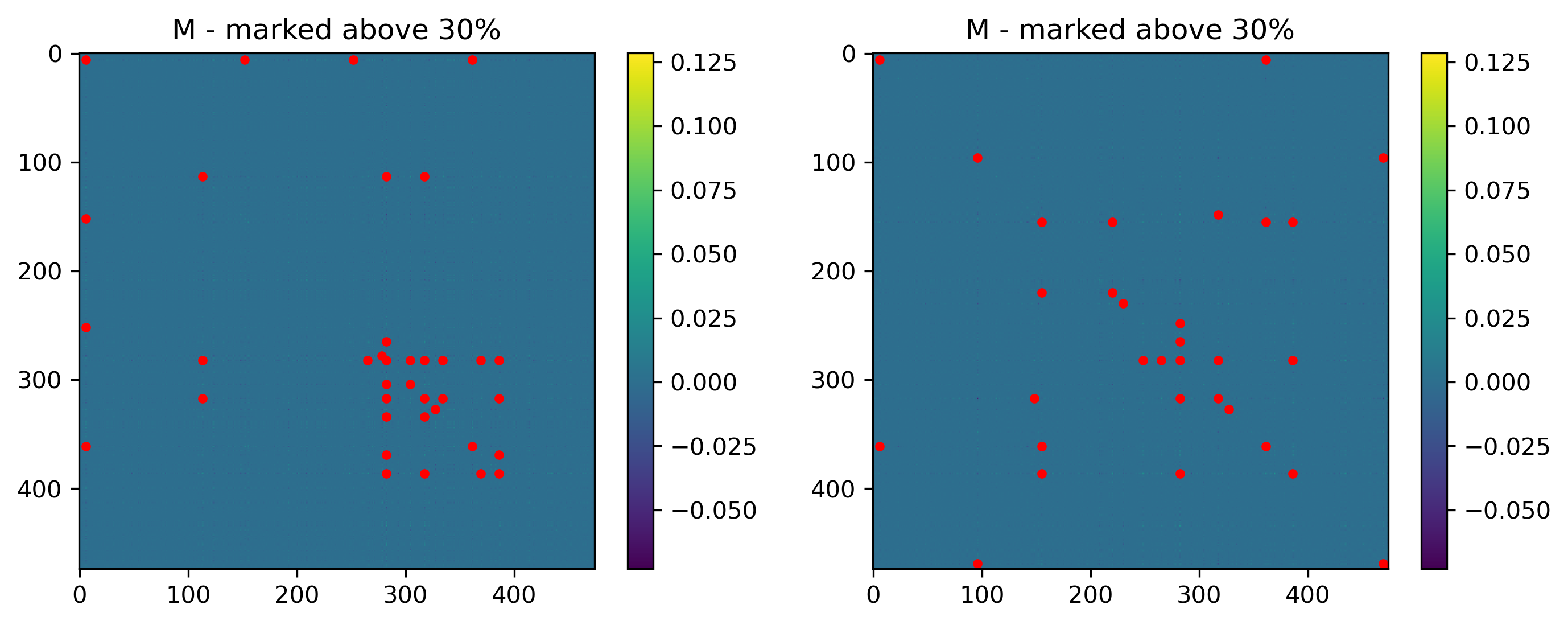}
\caption{Optimized metric matrix for various training sets. Red dots show entries with larger values than 30\% of the maximum.}
\label{fig:opt_M}
\end{figure}

Fig.\;\ref{fig:opt_M} shows visualizations of two example metric matrices in our method for different training sets. 
All entries larger than 30\% of the maximum entry are marked by a red dot.
Since the training sets are randomly constructed, different combinations of training nodes lead to different optimized metric matrices.
We observe that only a few entries (features) along the diagonal contributed substantially to label prediction. 

\begin{figure}[h]
\centering
\captionsetup{justification=centering,margin=0.1cm}
\includegraphics[width=0.45\textwidth]{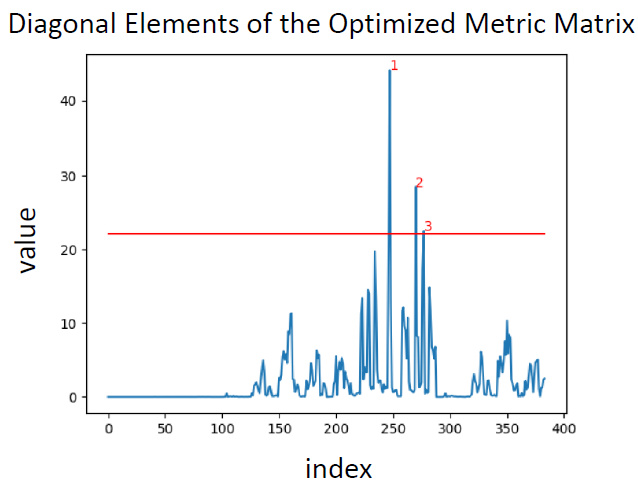}
\caption{Diagonal elements of an optimized metric matrix.}
\label{fig:diagonal-M-3dsift}
\end{figure}

These dominant features can be extracted and visualized. 
For example, Fig.\;\ref{fig:diagonal-M-3dsift} shows the diagonal entries of an optimized metric matrix using 3D-SIFT features extracted from the fish movie.
We can separate the dominant features via thresholding.
The red line in Fig.\;\ref{fig:diagonal-M-3dsift} shows an example of a threshold set at $50\%$ of the maximum entry.
How these relevant features are correlated to each other are encoded in off-diagonal terms.

The interpretability of the features themselves varies among different algorithms.
For example, 3D-SIFT is more interpretable than CNN-based features like Slowfast and SOE-Net.
Each element of the feature vector obtained by 3D-SIFT unambiguously corresponds to the gradients of a specific area of the frame batch in a specific direction.
This area is determined by the 3D-SIFT keypoint and the location of each element relative to the keypoint.

\begin{figure}[h]
\centering
\captionsetup{justification=centering,margin=0.1cm}
\includegraphics[width=0.49\textwidth]{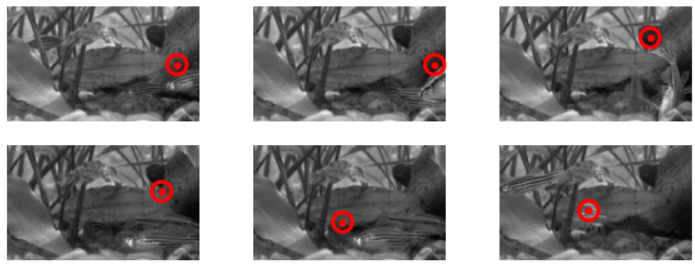}
\caption{3D-SIFT keypoints on the fish movie frames}
\label{fig:sift3d-kp}
\end{figure}

There may be any number of 3D-SIFT keypoints found in a frame batch. 
Nevertheless, all feature vectors should have the same length in our model; therefore, we need to select a fixed number of keypoints in each frame batch and disregard the others.
Many keypoint selection strategies can be adopted. One example, which is employed in our experiments, is selecting the keypoint closest to the center of the frame batch. 
To illustrate, Fig.\;\ref{fig:sift3d-kp} shows the selected 3D-SIFT keypoint on several frame batches from the fish movie.
Selecting the most central keypoint is a rudimentary provision to cover the largest possible area by the keypoint's descriptor.

\begin{figure}[h]
\centering
\captionsetup{justification=centering,margin=0.1cm}
\includegraphics[width=0.48\textwidth]{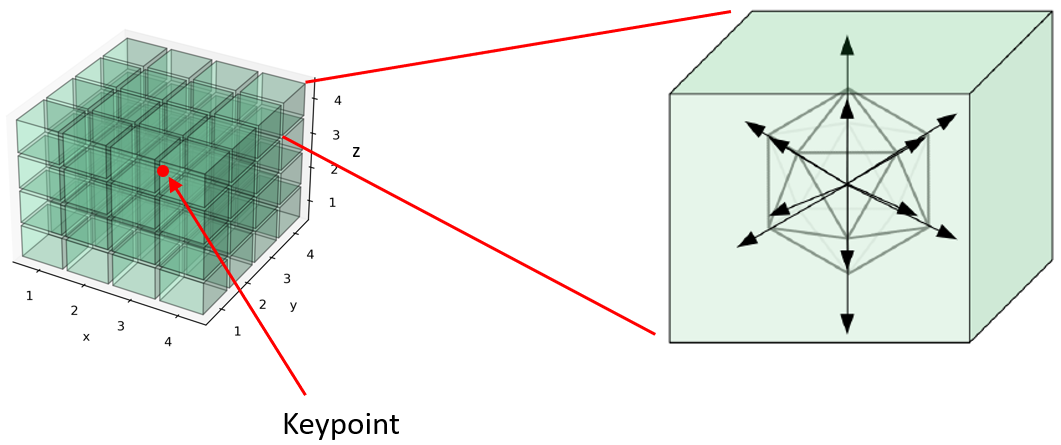}
\caption{The subregions around the 3D-SIFT keypoint and the icosahedron whose vertices correspond to histogram bins created in each subregion.}
\label{fig:3dsift-subregions}
\end{figure}

The keypoint descriptor is computed as follows.
A neighborhood around the keypoint is considered and divided into $64$ subregions. Local gradients are computed in each subregion and a histogram is constructed which shows the cumulative projections of the gradients in $12$ directions in the 3D space. 
Each direction corresponds to a vertex of a regular icosahedron.
To compute the cumulative gradient projections in each direction, only those local gradients are considered that intersect with at least one of the faces comprising the vertex in question.
Fig.\;\ref{fig:3dsift-subregions} shows the $64$ subregions and the icosahedron in one of the subregions.

\begin{figure}[h]
\centering
\captionsetup{justification=centering,margin=0.1cm}
\begin{subfigure}{.23\textwidth}
    \centering
    \includegraphics[width=\linewidth]{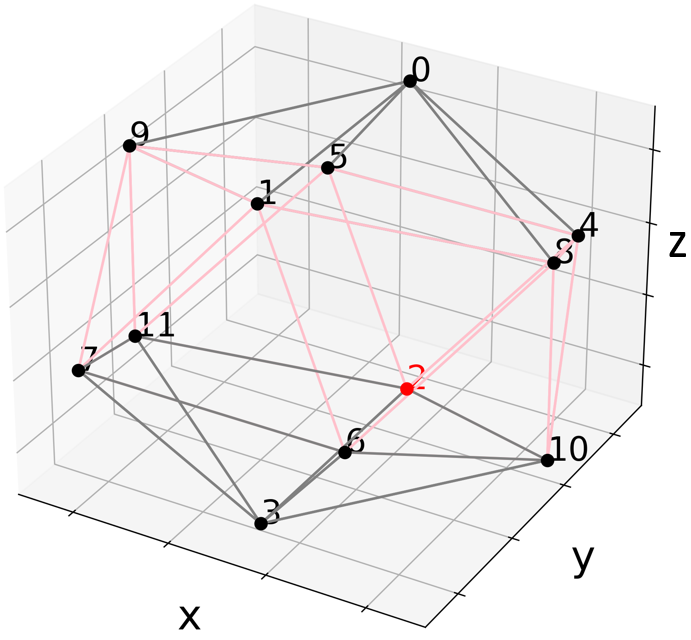}
    \caption{}
    \label{fig:icosahedron}
\end{subfigure}
\begin{subfigure}{.23\textwidth}
    \centering
    \includegraphics[width=\linewidth]{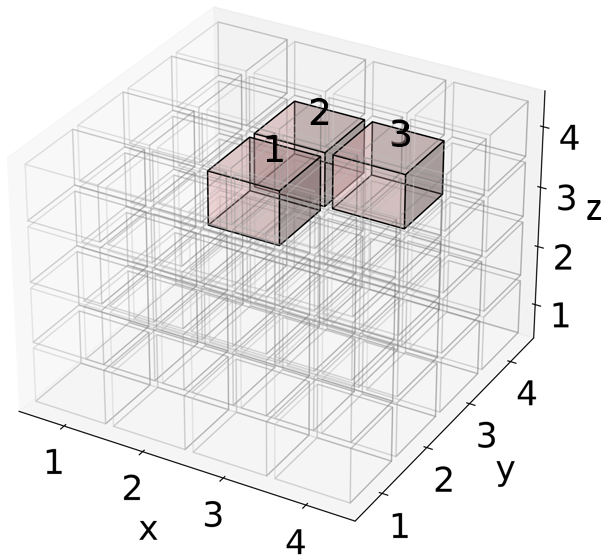}
    \caption{}
    \label{fig:3dsift-high-subregions}
\end{subfigure}
\caption{(a) Each element of a 3D-SIFT histogram corresponds to one of the $12$ vertices of a regular icosahedron. (b) Locations of the most dominant features relative to the keypoint for 3D-SIFT.}
\label{fig:top-subregions-icos}
\end{figure}

The subregions are numbered based on their Cartesian coordinates with respect to the keypoint. 
Similarly, the vertices of the icosahedron are also numbered as in Fig\;\ref{fig:icosahedron}.
The $64$ histograms, each consisting of $12$ bins, are concatenated to form the keypoint descriptor which we assign to the frame batch as its feature vector. In other words, the feature vector has \(12 \times 64 = 768\) dimensions.
Consequently, each element of a 3D-SIFT feature vector can be described by two numbers: one number determining its corresponding subregion relative to the keypoint, and one number depicting its corresponding vertex in the icosahedron (i.e. the general direction of the gradients).

Given this knowledge of what each element in a 3D-SIFT feature vector means, we can generate more detailed interpretations of the prediction process.
For instance, using the diagonal elements of \(\M^*\) from Fig.\;\ref{fig:diagonal-M-3dsift}, we can illustrate the locations of dominant features relative to the keypoint as in Fig.\;\ref{fig:3dsift-high-subregions}.

Furthermore, we can describe the dominant features with their direction and location relative to the keypoint.
For example, we can describe the most dominant feature in Fig.\;\ref{fig:diagonal-M-3dsift} as ``the cumulative projections of local gradients in the general direction of \((\theta = 148.28^o, \phi = 90^o)\) 
in spherical coordinates, computed in a subregion above the keypoint and to its northwest.''
This long description can be summarized by two numbers: \((2, 41)\). The first number (i.e. $2$) notes the corresponding icosahedron vertex of the most dominant feature. This vertex is marked in red in Fig\;\ref{fig:icosahedron}. The second number (i.e. $41$) depicts the corresponding subregion (i.e. location) of the most dominant feature. This subregion is marked by number $1$ and pink color in Fig\;\ref{fig:3dsift-high-subregions}.
The axes of Fig.\;\ref{fig:top-subregions-icos} are defined based on the keypoint's orientation which is known.

\begin{figure}[h]
\centering
\captionsetup{justification=centering,margin=0.1cm}
\includegraphics[width=0.48\textwidth]{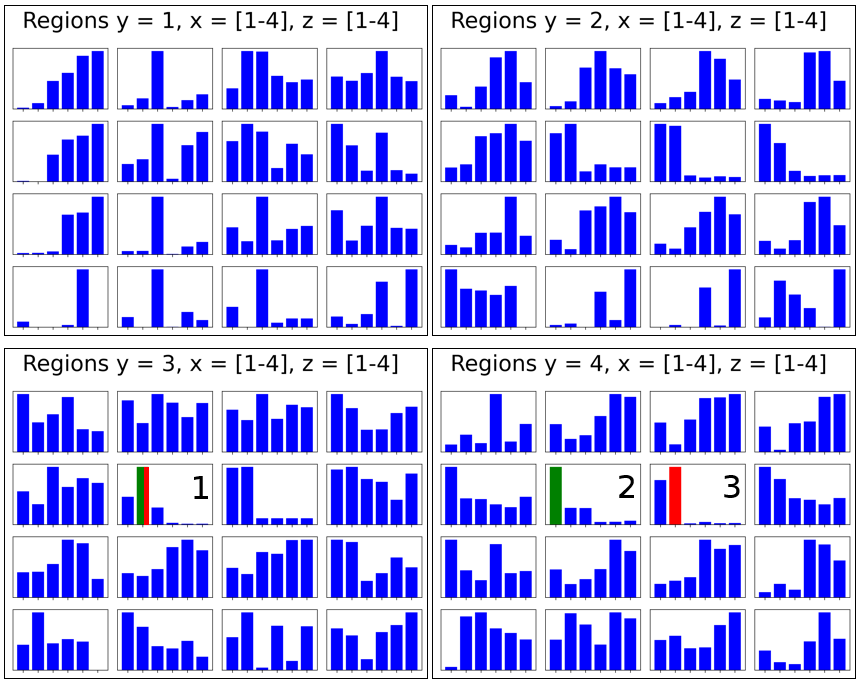}
\caption{Diagonal elements of \(\M^*\) visualized as histograms inside subregions around the 3D-SIFT keypoint. Subregions labeled $1$, $2$ and $3$ are the same subregions as in Fig.\;\ref{fig:3dsift-high-subregions}. Each bin in a histogram represents a specific direction in 3D space.}
\label{fig:histogram-M-3dsift}
\end{figure}

Fig.\;\ref{fig:histogram-M-3dsift} visualizes the diagonal elements of \(\M^*\) as histograms in $64$ subregions around the 3D-SIFT keypoint; thus, the contribution scores of all features are illustrated in one figure.
The subregions are distinguished by their coordinates \(x=\{1,2,3,4\}, y=\{1,2,3,4\}, z=\{1,2,3,4\}\).
We can locate the most dominant feature in Fig.\;\ref{fig:diagonal-M-3dsift} on this illustration as well; it corresponds to the $2$nd bin of the $42$nd histogram (\(x=2,y=3,z=3\)). This feature is distinguished by the color half-red half-green in Fig.\;\ref{fig:histogram-M-3dsift}.
Each bin in a histogram corresponds to an icosahedron vertex. The histograms in Fig.\;\ref{fig:histogram-M-3dsift} have $6$ bins instead of $12$; the reason is that the 3D-SIFT feature vectors were subsampled with a rate of $2:1$ to cut the time complexity into a quarter of its original value.

In addition to single features, we can illustrate \textit{pairs} of features whose corresponding non-diagonal element in \(\M^*\) have the largest values. 
Two such pairs are illustrated in Fig.\;\ref{fig:histogram-M-3dsift} with red and green colors.
The red pair correspond to the highest non-diagonal element, and the green pair correspond to the second highest.
A large non-diagonal element of \(\M^*\) means that the correlation between the two features is highly informative for label prediction.

\section{Conclusion} 
\label{sec:conclude}
To extract knowledge from ganglion cell responses to visual stimuli, we propose an interpretable graph-based binary classifier learning framework, where a positive semi-definite (PSD) metric matrix $\M$ is optimized to determine feature distances between nodes representing visual events.
Using a new objective called graph-based large margin nearest neighbor (GLMNN), we formulate a semi-definite programming (SDP) problem to optimize $\M$, which is efficiently approximated via an adoption of Gershgorin disc perfect alignment (GDPA) linearization. 
Given an optimized metric matrix $\M$, one can directly identify features substantially contributed to label classification by locating large-magnitude diagonal terms.
Experimental results show that our graph-based classifier is competitive with state-of-the-art classification schemes, while offering a level of interpretability. 

For future work, our model can be employed to predict the responses of individual ganglion cells, various physiological groups of ganglion cells, or even retinal responses of other organisms such as mice and primates \cite{rgctypes_mice, shlens2006structure}.
Analyzing the optimized model on these datasets can reveal further insight into the retinal computation.

\section{Acknowledgement}
We thank Benjamin D. Hoshal and Stephanie E. Palmer of the University of Chicago for constructive discussions.


\begin{scriptsize}
\bibliographystyle{ieeetr}
\bibliography{ref}
\end{scriptsize}

\end{document}